\documentclass[sn-basic,Numbered,smallcondensed]{sn-jnl}

\usepackage{graphicx}%
\usepackage{multirow}%
\usepackage{amsmath,amssymb,amsfonts}%
\usepackage{amsthm}%
\usepackage{mathrsfs}%
\usepackage[title]{appendix}%
\usepackage{xcolor}%
\usepackage{textcomp}%
\usepackage{manyfoot}%
\usepackage{booktabs}%
\usepackage{algorithm}%
\usepackage{algorithmicx}%
\usepackage{algpseudocode}%
\usepackage{listings}%

\theoremstyle{thmstyleone}%
\theoremstyle{thmstyletwo}%

\theoremstyle{thmstylethree}%

\begin{document}

\title[Overlooked and Underutilized: Prototypes are the Missing Piece in Tabular XAI]{Overlooked and Underutilized: Prototypes are the Missing Piece in Tabular XAI}


\author*{\fnm{Jacek} \sur{Karolczak}}\email{jkarolczak@cs.put.poznan.pl} 

\author{\fnm{Jerzy} \sur{Stefanowski}}\email{jstefanowski@cs.put.poznan.pl} 

\affil[]{\orgdiv{Institute of Computing Science}, \orgname{Poznan University of Technology}, \orgaddress{\street{ul. Piotrowo 2}, \city{Poznań}, \postcode{61-138}, \country{Poland}}}

\abstract{Explainable AI (XAI) is frequently applied to high-stakes tabular data, yet the field relies heavily on feature attribution methods. While these approaches offer technical transparency, evaluating dozens of isolated features demands excessive cognitive effort from non-expert users. In this paper, we argue that prototypes remain overlooked and underutilized, and should be integrated as a complementary, intuitive gateway to auditing complex machine learning models. Although entire exemplars function as an intuitive baseline for prototype-based explanations, we highlight prototypical parts as particularly valuable for tabular data. By grouping related features into bounded subspaces, prototypical parts technically embody human-understandable concepts, replacing isolated features with cohesive semantic units. We support this integration through two primary lenses. Cognitively, these parts align with Rosch’s categorization theory and Kahneman’s System 1 thinking to naturally reduce cognitive load for humans. Then, we introduce a Geometric Validity Triad to conceptualize why prototypes validate the data manifold---a critical dimension often neglected by existing methods. We contend that making tabular XAI genuinely comprehensible requires shifting toward composite explanation pipelines that pair intuitive similarity-based reasoning with rigorous feature-level auditing. The paper concludes by outlining open challenges of prototype-based methods and discussing potential solutions to guide future research.}



\begingroup
\renewcommand{\thefootnote}{}%
\footnotetext{\textbf{Acknowledgments:} This work was funded by the National Science Centre, Poland, under the OPUS programme (grant number 2023/51/B/ST6/00545).}
\endgroup

\maketitle

\begin{center}
{
\small
\textit{This is the original pre-peer-review version. The manuscript has been accepted for publication in the Journal of Intelligent Information Systems (Springer Nature).}
}
\end{center}

\section{Introduction}
\label{sec:introduction}
The rapid development of complex modern \emph{machine learning} (ML) systems has forced a resurgence of interest in explaining their decision-making processes. This renewed focus on \emph{Explainable AI (XAI)} was marked over 10 years ago, which galvanized the effort to \emph{open the black box}~\citep{riberio2016lime}. Since then, the XAI community has proposed a vast array of methods, many of which remain rooted in the principles of symbolic AI~\citep{Bodria2023}. However, this rapid expansion has created a landscape saturated with techniques of varying practical utility, often lacking rigorous comparative studies regarding their specific areas of competence~\citep{moshkovitz2026position}. 

Most of the recent progress in XAI has been catalyzed by the rise of deep neural networks in Computer Vision (CV) and Natural Language Processing (NLP). In both domains, explanation methods were often developed with a focus on  debugging, serving the needs of model developers~\citep{biecek2024positionexplainquestionjustify}.

While concept-based explanations were widely adopted in CV---exemplified by architectures like ProtoPNet~\citep{chen2019protpnet} and methods like Concept Relevance Propagation~\citep{achtibat2023crp}---research in the tabular domain has historically neglected the development of comparable human-understandable concept-based explanations.

However, explainability for tabular ML models remains a practical and legal necessity~\citep{Bodria2023,longo2024manifesto}, especially in high-stakes sectors such as healthcare~\citep{chen2024sepsis}, finance~\citep{liao2020bankingexample}, and government~\citep{borison2022tabreview}. Unlike images or text, tabular data lacks spatial smoothness and sequential continuity. Consequently, tabular ML systems are specifically designed to handle noisy columns, high-cardinality categorical data, and sharp piecewise decision boundaries, often achieving high efficiency on smaller datasets with minimal tuning. This distinct structure makes them well-suited for targeted post-hoc methods~\cite{moshkovitz2026position}.

Hence, the tabular XAI landscape is characterized by a heavy reliance on \emph{Feature Importance} (FI) methods, such as SHAP~\citep{lundberg2017shap} or LIME~\citep{riberio2016lime}, leading to a fragmented practice where other approaches are less frequently explored~\citep{Bodria2023,moshkovitz2026position}. While FI has become a very popular approach, it also creates barriers for lay users. As human-centered evaluations emphasize, algorithmic transparency does not inherently equate to comprehensibility~\citep{lopes2022xaievaluation}. The user must understand the FI's mathematical assumptions---a requirement that often renders the explanation inaccessible in practice~\citep{rudin2019explaining}. Furthermore, FI methods are sensitive to feature representation; trivial encoding changes (e.g., discretization) can drastically alter explanations~\citep{hwang2025shapsensitive}.

Furthermore, several empirical studies have increasingly exposed operational failures in how humans interact with explainability based on feature importance. For instance,~\citep{poursabzi2021manipulating} demonstrated that presenting users with clear, transparent feature importance vectors paradoxically impaired their ability to detect erroneous models compared to those using a black-box model. Similarly, experiments in~\citep{bansal2021doesthewhole} revealed that adding feature attribution failed to improve human-AI team performance; instead, it acted as an anchor that induced users to blindly accept erroneous predictions. These examples underscore a structural deficit in current XAI: an explanation does not automatically yield comprehension or critical auditing capacity~\citep{bansal2021doesthewhole,poursabzi2021manipulating}.

Consequently, symbolic methods such as \emph{rules}~\citep{riberio2018anchor} and \emph{counterfactuals}~\citep{guidotti2022counterfactuals} offer a transparency that abstract feature vectors lack. However, while highly interpretable, navigating extensive rule sets remains cognitively demanding, and interpreting complex counterfactuals can prove equally challenging for lay users.. In high-stakes decisions---ranging from credit lending~\citep{liao2020bankingexample} to medical diagnostics~\citep{chen2024sepsis}---the inability to grasp the model's rationale at a glance can hinder informed decision-making and erode trust.

To complement these methods, we highlight \emph{prototypes} as the missing piece in tabular XAI---an explanation modality that remains overlooked and underutilized. As we thoroughly discuss in Section~\ref{sec:cognitive}, prototypes naturally align with human reasoning by anchoring abstract categories to concrete examples rather than complex models or additive weights~\citep{rosch1975cognitive}. When properly structured into restricted feature subspaces that represent clear concepts, this alignment significantly reduces cognitive load, making prototypes  well-suited for rapid skim analysis and the bulk screening of predictions. Rather than overwhelming the user with exhaustive technical details upfront, they facilitate effortless gist extraction. This allows for quick plausibility checks that can organically trigger a need for deeper understanding, prompting the user to transition into more in-depth, cognitively taxing investigations---such as utilizing more advanced methods---only when a specific prediction needs further investigation.

Throughout this paper, unless specified otherwise, we use \emph{prototypes} as an umbrella term covering both:
\begin{itemize}
    \item \textbf{Entire Exemplars:} Complete instances from the dataset that serve as central representatives of a class, maximizing within-class similarity and between-class separation. These are typically identified via post-hoc methods to explain existing models~\citep{molnar2022}. 
    \item \textbf{Prototypical Parts:} Interpretable feature subspaces corresponding to human-understandable concepts that serve as the fundamental patterns defining a given class~\citep{chen2019protpnet}.
\end{itemize}

We unify these partly distinct forms under one definition because they both leverage similarity-based reasoning, diverging from other  methods. Prototypes can function as \emph{local} (explaining the decision of ML model for a single instance) and \emph{global} (approximating the overall internal logic of the model) explanations and are applicable in both ante-hoc (inherent to the model decision-making process) and post-hoc (explaining trained models) settings \citep{molnar2022}.

While entire exemplars serve as a good intuitive baseline, we argue that \emph{prototypical parts} are particularly interesting for addressing the structural complexity of tabular data. Although the generation of whole-instance prototypes was established in classical ML~\citep{biehl2016prototype}, the approach was largely neglected before its full potential could be realized. Yet the extraction of prototypical parts---common in computer vision---has not yet been successfully implemented in tabular data~\citep{narayanan2024geo}, primarily because the lack of spatial ordering and the presence of heterogeneous features demand entirely novel architectural designs. Consequently, we strongly advocate for prioritizing prototypical parts in tabular XAI (also posing the need to build new neural networks), promoting a necessary shift from matching full data rows to reasoning via compositional feature subspaces.

The necessity for methodological pluralism is underscored by the diversity of humans. Explanations are only valuable if they are understandable and actionable, and different stakeholders have vastly different needs~\citep{moshkovitz2026position}. Consequently, a method suitable for a data scientist debugging a model may fail for a lay user making a high-stakes decision without a background in ML~\citep{doshivelez2017rigorous}. 

\textbf{We contend that making tabular XAI genuinely comprehensible to lay users requires the research community to better integrate prototypes as a complementary explanation method that connects intuitive reasoning with fine-grained model understanding—a shift that necessitates solving several open challenges outlined in this paper.}

The remainder of this paper is organized as follows. Sections II--IV substantiate our core arguments by discussing the cognitive foundations of prototype-based reasoning, introducing the Geometric Validity Triad, and defining concepts within the tabular domain. Section V outlines the critical open challenges facing prototype-based methods---spanning evaluation, algorithmic advances, human-computer interaction, and privacy---and proposes specific future directions to guide subsequent research. Finally, Section VI provides concluding remarks on the necessary shift toward inherently interpretable, tabular-native architectures.

Notably, we do not propose a novel algorithmic explanation method. Instead, we analyze the cognitive and geometric limitations of current tabular XAI, propose integrating prototypes (and, in particular, prototype parts) as an intuitive gateway to deeper understanding, and outline the critical open challenges that must be resolved to realize this vision.

\section{The Cognitive Primacy of Prototypes}
\label{sec:cognitive}

The transition to the XAI field requires moving from engineering-centric transparency toward modalities that actively integrate cognitive science discoveries to align with the human cognitive model~\citep{longo2024manifesto,lopes2022xaievaluation}. Current XAI methods often assume an idealized user capable of high-load analytical processing~\citep{cau2025exploring,jong2025cognitive}. We argue that explanation efficacy depends on cognitive fit. This position is synthesized from a review of well-established cognitive science literature~\citep{rosch1975cognitive,kahneman2011thinking} and recent interdisciplinary reviews~\citep{longo2024manifesto}, which collectively point to similarity-based reasoning as the preferred mechanism for human comprehension~\citep{davoodi2023parthuman}. By grounding tabular XAI in these cognitive frameworks, we demonstrate that prototypes are not merely an alternative visualization but the cognitive path of least resistance, particularly for users without a background in ML.

\subsection{Rosch's Categorization Theory}
While machine learning classifiers typically operate by defining discriminatory boundaries (e.g., hyperplanes or decision tree thresholds) in high-dimensional space, users do not naturally process categories through such separation. According to Rosch’s categorization theory, humans classify objects by comparing them to a \emph{mental centroid}---a representative exemplar that encapsulates the central tendency of a category~\citep{rosch1975cognitive}.

This dissonance between model and user logic creates an interpretability gap. Consequently, XAI methods that present decision boundaries impose a translation tax on the user, requiring the reconstruction of a mental image from abstract constraints. Prototype-based explanations bridge this gap by mimicking the natural cognitive mechanism of similarity-based reasoning. For example, a human-centric study~\citep{davoodi2023parthuman} provides empirical evidence that presenting prototypical parts learned by classifiers truly enhances interpretability.

The user often employs the \emph{representativeness heuristic}---inherently evaluating likelihood by comparing a specific instance against a pre-existing mental prototype or stereotype~\citep{kahneman2011thinking}. The representativeness heuristic is particularly dominant in high-stakes expert domains. 

For instance, clinical reasoning is fundamentally driven by \emph{illness scripts}---mental prototypes of typical disease presentations~\citep{greengrass2026clinicalreasoning}. Without explicit prototypes, the user is forced to simulate this anchor internally, often relying on biased assumptions~\citep{byrne2023goodexplanations}. Providing an actual prototype mitigates this by replacing subjective mental images with a concrete anchor rooted in the model's true data distribution. Ultimately, as \citep{hoffman2023measures} notes, users do not evaluate pure mathematical weights; they map explanations onto internal schemas, and the degree of this alignment dictates their trust in the system.

\subsection{Bridging System 1 and System 2 Thinking}
The utility of explanations can be further analyzed through Kahneman’s dual-process theory~\citep{kahneman2011thinking}: \textbf{System 1} (fast, intuitive processing, such as instantly recognizing a favorite brand by its packaging color) and \textbf{System 2} (slow, analytical computation, such as stopping to calculate the price-per-ounce to find the best discount).

Standard tabular methods, such as feature importance or counterfactuals, inherently demand System 2 processing because they require the user to perform mental arithmetic and weigh competing features. For non-expert users, this cognitive burden is often prohibitive. Explanations demanding heavy System 2 engagement paradoxically cause users to revert to uncritical System 1 heuristics~\citep{jong2025cognitive}. 

This phenomenon manifests as the \emph{Illusion of Explanatory Depth} in XAI~\citep{chromik2021illusion}. User studies evaluating additive feature explanations reveal that while mathematical outputs increase self-reported confidence, they fail to improve objective understanding. Because feature-based outputs appear rigorous, users experience a false sense of comprehension and prematurely suspend critical auditing. Consequently, these explanations often act as persuasive anchors rather than diagnostic tools, inducing blind trust in both lay users~\citep{bansal2021doesthewhole} and data scientists~\citep{kaur2020interpretinginterpretability}.

In high-stakes domains, overcoming these cognitive limitations requires AI systems to align directly with human heuristics~\citep{greengrass2026clinicalreasoning}. Prototype-based reasoning leverages the strengths of System 1 because identifying a familiar, bounded conceptual part is an immediate and intuitive process. By presenting a structurally coherent prototypical part, we allow the user to grasp the essence of the decision via System 1, reserving their limited System 2 resources for critical auditing only when anomalies are detected.

\subsection{Metacognitive Monitoring}
The goal of XAI 2.0 is not merely reassurance but active understanding~\citep{longo2024manifesto}. Recent empirical studies in AI-assisted decision-making emphasize that standard explanations often fail to trigger this understanding, instead causing users to overrely on the AI. As one potential solution, Bu\c{c}inca et al.~\citep{buccinca2021totrustortothink} demonstrated that this overreliance can be mitigated using \emph{cognitive forcing functions}---interventions designed to disrupt fast, intuitive reasoning and force analytical engagement.

We argue that prototypes naturally act as these visual cognitive forcing functions. Because the user can intuitively compare the AI's extracted concepts---represented as a prototype---against the user's internal mental prototype, a semantic mismatch creates a strong cognitive divergence signal. This signal successfully disrupts intuitive reasoning and forces a shift to critical analysis. This mechanism aligns with the assertion by \citep{biecek2024positionexplainquestionjustify} that the purpose of explanation is to prompt the user to question the decision rather than merely justifying a prediction. Thus, prototypes facilitate a safe fail-over from System 1 to System 2 when the model's reasoning conflicts with user intuition.

\subsection{Cognitive Load and Information Chunking}
Current optimization-centric XAI typically presents maximization interfaces (e.g., ranked lists of dozens of features). This one-size-fits-all approach violates the bounded rationality framework, where decision-making is driven by \textit{satisficing}---searching for the first satisfactory hypothesis rather than computing an optimal answer~\citep{kahneman2011thinking}. 

While entire exemplars offer a more comprehensible baseline, they still induce cognitive overload when scaled to high-dimensional tabular data. Unlike an image, where a million pixels are instantly compressed by the human visual cortex into a single recognizable concept, a user cannot instantly perceive a 500-column tabular row. Instead, users must sequentially read and evaluate each variable, rapidly exceeding the $7 \pm 2$ limit of working memory~\citep{miller1956seven}. Consequently, lacking structural context, users fail to accurately anticipate model behavior~\citep{lavellehill2025explainable}.

Prototypical parts leverage the psychological mechanism of \emph{information chunking}. Since human reasoning is inherently selective and contrastive~\citep{miller2019explanation}, presenting a concept as a restricted feature subspace creates a single cognitive unit, instantly highlighting why one outcome occurred over another.

By isolating the prediction mechanism to this restricted subspace of just a few interacting features, the user recognizes the correlation structure as a single, coherent concept. This prevents cognitive overload, making prototypical parts a viable modality for high-dimensional tabular data.

The underlying psychological preference for this \emph{similarity-based reasoning} is corroborated by empirical evidence. A comprehensive user study in~\citep{jeyakumar2020howcaniexplain} demonstrated a consistent human preference for and better comprehension of explanations-by-examples over feature attribution methods across diverse data modalities, such as images, text, and audio. We argue that developing prototypical parts is the necessary algorithmic step to successfully translate this proven cognitive preference into the high-dimensional tabular domain.

\section{The Geometric Validity Triad}
\label{sec:triad}

The existing discourse in XAI often seeks a single, gold-standard solution for interpretability. However, because explanations must simultaneously align with both human cognition and the model's internal logic, we argue that such a singular approach is structurally insufficient. Instead, robust global model validation requires a multi-view perspective. To operationalize this, we propose a \emph{Geometric Validity Triad} that categorizes methods based on whether they validate the \textbf{Axes} (features), the \textbf{Decision Boundary} (decision limits), or the \textbf{Manifold} (data density).

\begin{figure}[t!]
    \centering
    \includegraphics[width=0.5\textwidth]{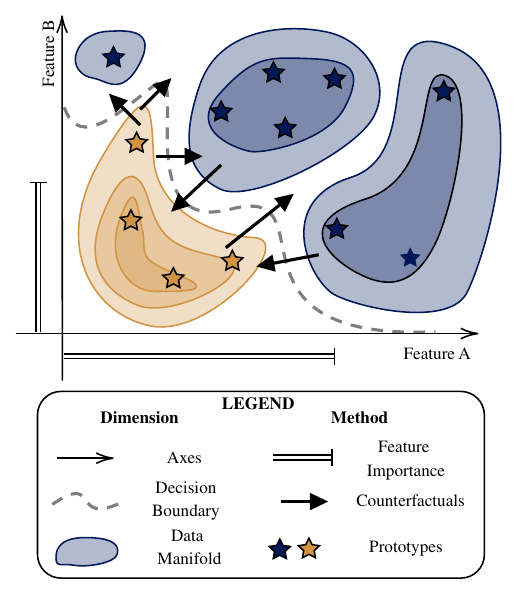}
    \caption{The Geometric Validity Triad: Visualizing the three layers of model validation. (1) Axes: Features define the coordinate system. (2) Boundary: Counterfactuals probe the decision boundary for tipping points. (3) Manifold: Prototypes validate the data density.}
    \label{fig:triad}
\end{figure}

The triad builds on the premise that robust global ML model validation requires a multi-view perspective to avoid the \emph{Rashomon effect}, where isolated explanations yield contradictory conclusions~\citep{baniecki2023rashomon}. To validate a model's logic, a data scientist must identify relevant features to define the space (Axes), delineate the precise tipping points of the prediction (Boundary), and finally verify that the explanations traverse populated regions of the feature space (Manifold). While high-NFC users may require traversing all three layers up to the abstract axes, the geometric foundation must always be anchored in the data manifold to ensure plausibility. The Geometric Validity Triad is visually depicted in Figure~\ref{fig:triad}.

\subsection{The Axes: Feature Importance}
\emph{Feature Importance (FI)} methods, such as SHAP~\citep{lundberg2017shap}, permutation importance, or others, operate by assigning scalar weights to input features.

In the Geometric Validity Triad, FI methods validate the Axes, ensuring the decision boundary is projected onto the most relevant feature dimensions.

This dimension provides feature attribution to establish the model's coordinate system, enabling diagnostics, sanity checks, and leakage detection (e.g., an \texttt{Index\_Column} driving predictions~\citep{lundberg2017shap}). However, validating the axes is often geometrically insufficient for explaining a specific decision. Knowing \emph{which} dimensions matter neither reveals the shape of the decision function nor the density of data within that space. As \citep{rudin2019explaining} argues, an importance vector explains the abstract mathematical coordinates but fails to justify the instance's position within a human-relatable narrative.

\subsection{The Decision Boundary: Counterfactuals}
\emph{Counterfactual} explanations operate by identifying the minimal perturbation required to alter the model's prediction~\citep{guidotti2022counterfactuals}. 

In the Geometric Validity Triad, counterfactual methods refer to showing actions towards the decision boundary.

One can view this as a form of cause-effect (what if) analysis. It moves beyond static axes to define the dynamic limits of the decision class, answering the question: \emph{What change has to be made to alter the decision?} This provides actionability and recourse, allowing domain experts to probe the stability of prediction~\citep{byrne2019counterfactuals}. However, boundary validation faces a critical limitation: the manifold problem. While a counterfactual explicitly indicates a new point beyond the decision boundary, this point may be located in an empty region (out-of-distribution). Basic methods for generating counterfactuals often optimize for the minimal distance and validity of changing the model prediction rather than data density, yielding perturbations (e.g., a 10-year-old with a PhD) that are valid regarding the decision but practically impossible to realize~\citep{upadhyay2025counter}. Therefore, recent approaches also propose plausible counterfactuals to locate generated points in regions densely populated by training instances~\citep{wielopolski2024plausible}.

\subsection{The Manifold: The Missing Link}

To complete the Geometric Validity Triad, we identify a critical gap in the current XAI landscape: while widely adopted tools excel at describing features (Axes) and decision tipping points (Boundaries), they may fail to validate the underlying distribution. Prototypes fill this void by confirming that the model is interpolating between known examples rather than extrapolating based on spurious correlations in sparse regions~\citep{longo2024manifesto}. This aligns with the natural heuristic of similarity, offering a sanity check that abstract weights cannot provide~\citep{cau2025exploring,rudin2019explaining}.

To bridge this gap, we advocate for prototype-based methods that explain decisions by referencing concrete data representations. These encompass both entire exemplars~\citep{kim2016criticizm} and prototypical parts~\citep{chen2019protpnet}.

Prototypes validate that the decision boundary is not merely an abstract hyperplane but is anchored in the manifold---in regions characterized by a high conditional probability density of observing instances from the given class. While entire exemplars validate the global joint distribution by providing a literal data point, prototypical parts validate the regional density, ensuring that the model relies on combinations of features that frequently co-occur in the training distribution rather than in sparsely populated regions of the subspace.

Furthermore, anchoring decisions in the data manifold naturally facilitates Out-of-Distribution (OOD) detection. In high-stakes domains, knowing when a model is guessing is as critical as understanding its reasoning. If a new instance falls far from all learned prototypical parts, it serves as an immediate structural warning that the input is out-of-distribution.

\subsection{Excluded Explanation Methods}
We exclude popular methods from the Geometric Validity Triad that rely on post-hoc surrogate models rather than directly exposing the model's internal geometry.

\subsubsection{Symbolic Explanations}
Decision trees and rule-based explanations are historically recognized as highly interpretable ML methods, offering a symbolic form of knowledge representation. They function both as interpretable global models and as local proxies approximating black-box decisions~\citep{molnar2022}.

However, despite recent advances (e.g.,~\citep{rudin2023rules}) advocating for their utility in high-stakes decision-making, we exclude them from the scope of the Geometric Validity Triad discussion. Their mode of generation and representation diverges significantly from the continuous geometry of Feature Importance, Counterfactuals, and Prototypes. From the perspective of geometric validity, trees or rule-based methods impose a hyper-cube discretization on the feature space, approximating complex, non-linear decision boundaries with coarse, axis-aligned conditional hyperplanes~\citep{tan2020treespace}. Furthermore, as global explanations, they often lead to deep trees or exhaustive rule lists that could be cognitively overwhelming.

\subsubsection{Explanations from LLMs}
Finally, we caution against the recent trend of using Large Language Models (LLMs) to generate narratives for tabular predictions. While LLMs produce fluent text, they frequently function as unfaithful approximations~\citep{wolff2025llmsreasontabulardata}. As highlighted by \citep{fang2024llm}, these models can generate plausible-sounding justifications that have no mathematical basis in the underlying model's logic. Recent studies further expose severe structural limitations: because LLMs process numerical data through text tokenization, they inherently discard critical geometric properties such as magnitude and spatial relationships~\citep{murugesan2026structural}. Consequently, the models often inject spurious correlations learned from linguistic pre-training rather than relying on the actual statistical distributions within the dataset. Furthermore, LLMs reasoning is vulnerable to format perturbations (e.g., transposing rows and columns or altering structural serialization), proving that LLMs lack genuine structural awareness of the data manifold~\citep{liu2024rethinking}. This creates a dangerous illusion of understanding, in which the user trusts the linguistic coherence---or unfaithful Chain-of-Thought narratives---rather than the true causal mechanism. 

\section{Defining Tabular Concepts via Prototypical Parts}

A fundamental limitation of most current tabular XAI is its reliance on raw input features as the atomic unit of explanation. We argue that user-centric explainability~\citep{longo2024manifesto} cannot be fully realized in the tabular domain until we move beyond atomic features toward disentangled representations composed of concepts. For instance, explaining a prediction via independent features such as \emph{age} and \emph{zip code} is cognitively less efficient than explaining it via a derived concept of \emph{socio-economic status}. To achieve this, we must first define the concept in tabular data.

\subsection{Concept: From Pixels to Feature Subspace}
In computer vision, a \emph{concept} is intuitively defined as a visual part or object, such as stripes, wheels, or beaks~\citep{chen2019protpnet}. In the tabular domain, we define a concept as a coherent interaction of specific feature-values within a feature subspace. Unlike an unbounded feature (e.g., \textit{Income}), a concept should be defined by specific value constraints (e.g., \textit{Financial instability: Income $\in [ \$15k,\,\$25 k]$ $\land$ Debt $>$ \$5k})~\citep{michalski1983subgroup}. In the context of modern XAI architectures, a \emph{prototypical part} serves as the technical realization of this abstract concept.

Crucially, we define such concepts as functionally coherent with respect to the model’s decision-making. 

A concept represents a grouping of features that the model utilizes as a single unit of reasoning to assign an instance to a given class, which is similar to definitions considered recently in neurosymbolic learning~\citep{silver2023roles}. This definition does not always require immediate human interpretability; rather, it enables global model validation.

Let us consider an illustrative example:  if a tabular neural network constructs a \emph{prototypical part} by isolating a bounded feature subspace---for example, $\text{Area} \geq 250\text{ m}^{2} \wedge \text{Distance\_to\_Center} \leq 2.0\text{ km} \wedge \text{Year\_Built} \geq 2020$---that part functions as the technical embodiment of a \emph{Luxury Property} concept. If this discovered bounded subspace aligns with domain expertise, it validates the model's internal rationale. Conversely, if an ML system aggregates a feature subspace that is functionally disjointed from a human perspective---analogous to an image classifier using background artifacts like car reflectors to characterize a bird species~\citep{baniecki2025cars}, or a tabular model mistakenly binding an arbitrary Applicant ID with a Timestamp ($\text{ID} \in [1000, 5000] \wedge \text{Time} > \text{12:00}$) to predict financial risk---it instantly exposes an unsound shortcut logic, allowing users to detect structural vulnerabilities and reject the model's reasoning, even if its overall predictive metrics remain empirically high.

While entire exemplars effectively anchor a prediction to a known case, prototypical parts are particularly interesting because they reveal the precise functional mechanisms driving that prediction. Ultimately, prototypical parts offer a highly valuable complementary approach for interpreting multidimensional tabular data.

Rather than overwhelming working memory with dozens of variables in isolation~\citep{miller1956seven}, prototypical parts isolate specific concepts within restricted feature subspaces. This naturally aligns with human reasoning and holistic pattern recognition~\citep{rosch1975cognitive,kahneman2011thinking}. Integrating this perspective helps bridge the interpretability gap, making tabular XAI significantly more intuitive.

\subsection{The Deficit of Tabular Concepts}

We characterize the current state of using XAI for tabular data as a missed opportunity for concept learning. Recall that early research in symbolic ML, especially within rule-based models, prioritized the discovery of concepts over simple instance classification~\citep{michalski1983subgroup}. Similarly, the field of Knowledge Discovery and Data Mining established approaches like Subgroup Discovery~\citep{lavrac2004subgroup} to identify interesting local patterns (groups of examples with characteristic descriptions). However, this methodological heritage has been overlooked in current XAI.

While some computer vision models learn to assemble decisions from parts (e.g., ProtoPNet~\citep{chen2019protpnet}), tabular XAI remains fixated on isolated feature attribution (e.g., SHAP~\citep{lundberg2017shap}). This ignores the semantic grouping---the joint contribution of several features---that constitutes the model's actual decision boundary. 

Crucially, it must be emphasized that transitioning prototypical parts from computer vision to the tabular domain is not a trivial algorithmic translation. In computer vision architectures, discovering parts is facilitated by the homogeneous nature of pixels and spatial locality; a concept is simply a spatial crop of a convolutional feature map. Conversely, tabular data exhibits no spatial ordering and is highly heterogeneous, mixing continuous magnitudes with discrete categorical variables. Furthermore, attempting to group features or calculate similarity within a raw, high-dimensional tabular space is mathematically unstable due to distance concentration. Consequently, one cannot simply append a prototype layer to a standard tabular network. Realizing prototypical parts in this domain requires fundamentally redesigning the representation space---shifting from spatial convolutions to isolating value-bound feature subspaces within a dense, lower-dimensional latent space---and completely redefining how internal distance metrics handle disparate data types.

Due to these severe architectural barriers, the number of attempts to propose prototypical-part neural networks remains remarkably low. Furthermore, the few existing methods often target highly specific tasks, such as survival analysis~\citep{kirpichenkoetal2025survival}, and specialized designs offer limited generalizability to other applications and domains:

\begin{itemize}
    \item \textbf{Manual Definition Dependency:} \citep{choi2022concept} requires users to define concepts by manually curating sets of positive and negative examples. This assumes perfect domain knowledge to construct valid example sets and prevents the automatic discovery of unknown or spurious concepts used by the model.
    \item \textbf{Shallow Feature Masking:} \citep{espinosa2023tabcbm} introduces TabCBM, which defines a concept as masking---a weighted selection of features. We argue that this definition is insufficient, as it is value-agnostic; it identifies which features matter, but not what value combinations (e.g., high vs. low) constitute a concept.
    \item \textbf{Domain Specificity:} \citep{karolczak2025medic} proposes a concept representation based on a combination of reference norms for laboratory testing, limiting its generalization to non-medical domains.
\end{itemize}

Consequently, tabular XAI currently lacks architectures capable of autonomously discovering generalizable prototypical parts. Moving beyond domain-specific heuristics to networks that natively learn coherent feature subspaces introduces several structural challenges, which we outline in the following section.

\section{Open Challenges and Future Directions}

To successfully apply prototypes  to tabular XAI systems, the field must resolve several gaps. In this section, we outline the open challenges for prototype-based tabular explanations, grouped into four primary areas: evaluation, algorithmic advances, human-computer interaction, and data privacy. For each domain, we define the \textit{Desired State} of the field, briefly review the \textit{Current State} to identify existing limitations, and propose a specific \textit{Future Direction} to guide subsequent research.

\subsection{Evaluation and Benchmarking}

\subsubsection{Insular Evaluation Standards}
\textbf{Desired State:} The evaluation of XAI methods relies on standardized, objective metrics that explicitly quantify desired properties, such as manifold coverage, alongside practical utility across diverse domains.

\noindent
\textbf{Current State:} Initial frameworks for evaluating prototypes, such as ProtoScore~\citep{monke2025framework}, prioritize discriminatory utility analysis over geometric validity. Formal, data-driven metrics to assess manifold coverage~\citep{karolczak2026prototypes} remain in their infancy and are rarely adopted as universal standards.

\noindent
\textbf{Future Direction:} Establish unified evaluation frameworks that seamlessly integrate mathematical rigor with human-centric utility. Researchers must prioritize standardizing formal geometric metrics, such as manifold coverage~\citep{karolczak2026prototypes}. Crucially, because evaluating structural sub-components like prototypical parts introduces \emph{unspecified explanation properties} that standard proxy measures cannot capture, the community must develop novel metrics tailored specifically to feature subspaces while concurrently demonstrating how these structurally sound prototypes explicitly improve expert tasks~\citep{doshivelez2017rigorous,moshkovitz2026position}.

\subsubsection{Lack of Ground Truth}
\textbf{Desired State:} Researchers validate the quality of discovered prototypical parts against universally accepted tabular benchmarks. In these datasets, ground-truth concepts are explicitly defined as known, value-bound feature subspaces, allowing for the objective measurement of how precisely a network isolates a valid prototype versus a spurious correlation.

\noindent
\textbf{Current State:} The field lacks tabular datasets with well defined ground-truth concepts, researchers must manually review the generated feature subspaces. However, attempting to visually inspect multidimensional tabular interactions introduces subjective uncertainty. Unlike in computer vision, where humans can reliably verify visual parts, this reliance on human intuition prevents automated benchmarking and completely obscures whether the model has isolated a valid structural pattern or simply memorized a spurious correlation~\citep{doshivelez2017rigorous}.

\noindent 
\textbf{Future Direction:} Construct synthetic tabular benchmarks where ground-truth concepts---mathematically defined as precise, value-bound feature subspaces---are explicitly embedded within the generative process. This establishes a much-needed tabular counterpart to the CUB-200 dataset~\citep{wah2022cub200}, enabling researchers to objectively measure a model's fidelity and structural disentanglement.

\subsubsection{Real-World Cognitive Utility}
\textbf{Desired State:} Evaluation frameworks rigorously measure human cognitive engagement and behavioral outcomes in high-stakes environments, proving that prototypes actively improve decision-making.

\noindent
\textbf{Current State:} Most evaluation protocols are designed solely by computer scientists. As recent literature highlights, while current evaluation heavily focuses on algorithmic faithfulness, genuine human evaluation remains rare~\citep{moshkovitz2026position}. They frequently rely on overly subjective user satisfaction metrics rather than better measuring actual task performance or cognitive impact~\citep{lopes2022xaievaluation,davoodi2023parthuman}. Theoretical claims regarding cognitive benefits~\citep{kahneman2011thinking,rosch1975cognitive} remain largely unproven in applied, high-stress settings.

\noindent
\textbf{Future Direction:} Establish deep collaborations with researchers from human-centered fields, such as psychology and cognitive science, to co-design evaluation frameworks. The field must explicitly shift towards evaluating the utility of explanations for specific, application-dedicated tasks (e.g., auditing or decision-making) rather than relying on abstract proxy measures~\citep{moshkovitz2026position}. These frameworks must move beyond simple user's satisfaction to measure constructive distrust---validating whether the explanation successfully triggers a switch to System 2 analysis during model errors~\citep{byrne2023goodexplanations}.

\subsection{Algorithmic and Architectural Advances}

\subsubsection{The Architectural Gap}
\textbf{Desired State:} The tabular machine learning ecosystem features natively interpretable, ante-hoc prototypical part architectures that deliver high predictive performance without relying on post-hoc approximations.

\noindent
\textbf{Current State:} Tabular XAI remains largely post-hoc~\cite{moshkovitz2026position}, dominated by surrogate models applied to standard network architectures like FT-Transformer or TabNet~\citep{borison2022tabreview}. Because tabular data lacks spatial neighborhoods, neural networks often struggle with irregular target functions and underperform compared to gradient-boosted decision trees~\citep{grinsztajn2022whydt,shwartz2022deeplearningisnot}. A few pioneering works attempt to integrate concepts natively (e.g., TabCBM~\citep{espinosa2023tabcbm} or the domain-specific MEDIC network~\citep{karolczak2025medic}), but they represent preliminary attempts rather than a standardized architectural family.

\noindent
\textbf{Future Direction:} Develop and benchmark prototypical part neural networks tailored specifically to the tabular domain. Building upon representation learning frameworks like PTaRL~\citep{ye2024ptarl}, future work must address the unique heterogeneity of tabular features. This requires formulating novel optimization constraints---such as strict latent space orthogonality and feature sparsity---to enforce disentanglement, alongside constructing discrete bottleneck layers or differentiable un-projection mappings that guarantee continuous latent prototypes can be reliably decoded back into exact, human-readable feature boundaries.

\subsubsection{Metric Sensitivity for Tabular Data}
\textbf{Desired State:} Similarity computations perfectly capture functional and semantic relationships within heterogeneous tabular data, seamlessly handling continuous and categorical features across varying scales.

\noindent
\textbf{Current State:} Standard distances fail to capture functional similarity~\citep{aggarwal2001surprising}. While advanced metrics like the Heterogeneous Value Difference Metric~\citep{wilson1997hvdm} mitigate some issues, they remain limited to basic input-space relationships. Recent measure-driven approaches successfully quantify underlying distributions~\citep{karolczak2026prototypes}, but they are designed solely for post-hoc evaluation rather than natively guiding deep networks.

\noindent
\textbf{Future Direction:} Adopt model-intrinsic similarity measures within tabular machine learning systems. While there are examples of successful formalization and adoption of intrinsic distance metrics for tree ensembles~\citep{tan2020treespace}, future research must adapt this paradigm to neural networks. This requires projecting heterogeneous inputs into a structurally aware latent space where distance is computed exclusively using the network's internal inference states (e.g., intermediate embeddings). This ensures that proximity strictly quantifies functional equivalence within the model's actual decision boundary, successfully bypassing input-space artifacts.

\subsubsection{Modeling Concept Complexity and High Dimensionality}
\textbf{Desired State:} Tabular machine learning systems inherently support hierarchical and compositional concept learning, dynamically compressing thousands of raw features into human-digestible, higher-level concepts.

\noindent
\textbf{Current State:} Current prototype methods assume concepts are flat and operate in lower-dimensional spaces. Processing thousands of raw features is impossible for human abilities~\citep{miller1956seven}, yet the field lacks  counterparts to classical conceptual clustering approaches (e.g., COBWEB~\citep{fisher1987knowledge}) that successfully model hierarchies.

\noindent
\textbf{Future Direction:} Investigate hierarchical concept learning and global-to-local feature selection. Future tabular networks should recursively combine lower-level feature interactions into higher-level abstract definitions (e.g., utilizing architectures like ProtoGate~\citep{jiang2024protogate}), allowing users to view aggregated concepts.

\subsubsection{Task Restriction and Data Stationarity}
\textbf{Desired State:} Prototype-based explanations for tabular data generalize across more complex predictive tasks, such as continuous regression and anomaly detection, while also dynamically adapting to concept drift in evolving data streams.

\noindent
\textbf{Current State:} This restrictiveness is a systemic limitation across most XAI methods. Whether utilizing feature attributions, counterfactuals, or prototypes across tabular, image, or text modalities, the vast majority of XAI methods are designed for simple classification. Adapting these methods for complex structured outputs---such as continuous regression~\citep{filho2023mpeer} or other semantic segmentation---remains a marginal area of research. Furthermore, standard XAI methods rely on  stationarity assumptions (static data and models). Because they are designed to explain a frozen model evaluated on a static dataset, they lack the mechanisms to incrementally  (online ) update explanations, rendering them obsolete when concept drift occurs in data streams or continual learning environments~\citep{karolczak2026prototypes}.

\noindent
\textbf{Future Direction:} Expand tabular prototype generation beyond static classification by engineering architectures capable of continuous adaptation. To tackle concept drift, tabular models should employ dynamic memory buffers and explicit prototype replay mechanisms, ensuring new structural patterns are assimilated without erasing previously learned concepts~\citep{yuan2024glance}. Moreover, dependencies between changes in data, models, and explanations in evolving systems should be studied \cite{stepka2026explaining}. Simultaneously, to support continuous targets and anomaly detection, researchers must shift the underlying objective functions away from discrete cross-entropy, instead optimizing prototypes as distance-based anchors distributed across a continuous target space~\citep{filho2023mpeer}.

\subsection{Human-AI Interaction and Cognitive Load}

\subsubsection{Cognitive Overload and Sparsity}
\textbf{Desired State:} Prototype methods present a strictly pruned, mathematically optimal set of exemplars that  captures the decision boundary without exceeding human working memory limits.

\noindent
\textbf{Current State:} High-performing methods generate an excessive number of prototypes to capture intra-class variance~\citep{chen2019protpnet,tan2020treespace}. Many architectures allocate a fixed number of prototypes per class, which scales linearly; however, tuning their number is not an easy task. Presenting too many prototypes to a lay user immediately exceeds working memory capacity and induces an overwhelming intrinsic cognitive load, actively defeating the explanation's purpose~\citep{miller1956seven}.

\noindent
\textbf{Future Direction:} Quantify the accuracy-sparsity trade-off through rigorous empirical human-subject studies. While vision architectures have developed algorithmic pruning to enforce prototype sparsity (e.g., ProtoTree~\citep{nauta2023pipnet}), future tabular research must formally evaluate how explanation size impacts human task performance and response time~\citep{lage2019evaluation}. Adopting this methodology is critical to establishing a tabular-specific counterpart to Miller's $7 \pm 2$ rule~\citep{miller1956seven}, ensuring networks minimize cognitive load without sacrificing fidelity.

\subsubsection{The Deficit of Interactive Analysis}
\textbf{Desired State:} XAI interfaces function as composite, interactive environments where users actively probe models, smoothly transitioning between intuitive prototype matching and fine-grained feature attribution. 

\noindent
\textbf{Current State:} The presentation layer is frequently treated as an afterthought. Explanations are rendered as static artifacts (e.g., flat tables), stripping away the user's ability to probe the model via \emph{what-if} manipulations. This ignores the established benefits of interactive explanatory model analysis~\citep{baniecki2023rashomon} and risks oversimplifying complex decision boundaries if prototypes are viewed in isolation~\citep{rudin2019explaining}.

\noindent
\textbf{Future Direction:} Integrate prototypes into a composite, interactive XAI framework alongside other XAI methods~\citep{longo2024manifesto}. Leverage interactive interfaces with inverse projections to help users visually audit semantic similarities and perturb inputs. Furthermore, interactive prototypes provide a stable form of algorithmic recourse. While a counterfactual provides the minimal perturbation to cross a decision boundary---often placing the user precariously close to the edge of rejection---a prototypical part represents a more robust, safe target destination. Future interfaces should allow users to map their current state against these prototypes to plan actionable, long-term recourse rather than minimal, potentially implausible shifts~\cite{wielopolski2024plausible}.

\subsection{Privacy and Security}

\subsubsection{Privacy Leaks from Entire Exemplars}
\textbf{Desired State:} Explanations provide high geometric fidelity and intuitive similarity matching without exposing any sensitive or personally identifiable information from the training distribution.

\noindent
\textbf{Current State:} Using real training instances as entire exemplar prototypes poses severe privacy risks under regulations like GDPR. In sensitive domains like healthcare or finance, highlighting actual patient records as global prototypes creates extreme vulnerabilities to membership inference and linkage attacks~\citep{wahdany2025privacy}.

\noindent
\textbf{Future Direction:} Develop generative methods for creating synthetic prototypes. These mechanisms must mathematically guarantee differential privacy while strictly maintaining distributional faithfulness to the original data manifold.

\section{Concluding Remarks}

The maturation of tabular XAI requires expanding how we present model logic to users. In this paper, we argue that prototype-based explainability, with particular emphasis on representing semantic concepts as prototypical parts, is still the missing piece in tabular data. While currently overlooked and underutilized, prototypes offer a complementary tool that naturally aligns with human reasoning. Framing prototypical parts as  restricted feature subspaces ensures they correspond strictly to human-understandable concepts. This conceptual grouping makes multidimensional tabular data substantially more intuitive alongside standard XAI methods.

Realizing this potential demands the development of new, inherently interpretable, tabular-native ML architectures. Consequently, newly proposed prototypical neural networks should be capable of mapping mixed categorical and continuous variables into structurally aware latent spaces. Future algorithmic designs must embed the autonomous discovery of an acceptable number of value-bound prototypical parts directly into the network training pipeline. This ensures that the discovered parts are intrinsic to the reasoning process, exactly reflecting the network's functional boundaries rather than relying on post-hoc approximations.

Finally, successfully integrating prototypical parts into the broader XAI ecosystem hinges on rigorous evaluation and new  human interactive design. The research should  move beyond  algorithm-centric proxy metrics and adopt frameworks that explicitly prioritize human-grounded evaluation. Rather than relying on subjective satisfaction surveys, future research must validate prototypical parts through specialized, human-in-the-loop studies that  measure cognitive load and task performance. 

\section*{Statements and Declarations}

\noindent
\textbf{Funding} This work was funded by the National Science Centre, Poland, under the OPUS programme (grant number 2023/51/B/ST6/00545).

\noindent
\textbf{Competing Interests} The authors declare that they have no financial or non-financial interests that are directly or indirectly related to the work submitted for publication.


\bibliography{bibliography.bib}

\end{document}